\pdfoutput=1

\documentclass[11pt]{article}

\usepackage[final]{acl}

\usepackage{times}
\usepackage{latexsym}

\usepackage[T1]{fontenc}

\usepackage[utf8]{inputenc}

\usepackage{microtype}

\usepackage{inconsolata}

\usepackage{graphicx}
\usepackage{tabularx}
\usepackage{float}
\usepackage{multirow}
\usepackage{booktabs}
\usepackage{colortbl}
\usepackage{bbding}
\usepackage{enumerate}
\usepackage{color}
\usepackage{amsthm,amsmath,amssymb}
\usepackage{mathrsfs}

\newcolumntype{C}{>{\centering\arraybackslash}X}

\title{SQLForge: Synthesizing Reliable and Diverse Data to Enhance Text-to-SQL Reasoning in LLMs}

\author{
  \textbf{Yu Guo\textnormal{\textsuperscript{1}}},
  \textbf{Dong Jin\textnormal{\textsuperscript{2}\footnotemark[1]}},
  \textbf{Shenghao Ye\textnormal{\textsuperscript{1}}},
  \textbf{Shuangwu Chen\textnormal{\textsuperscript{1}\footnotemark[1]}},
  \textbf{Jian Yang\textnormal{\textsuperscript{1}}},
  \textbf{Xiaobin Tan\textnormal{\textsuperscript{1}}}
\\
  \textsuperscript{1}University of Science and Technology of China
\\
  \textsuperscript{2}Institute of Artificial Intelligence, Hefei Comprehensive National Science Center
\\
  \texttt{\{yukariguo, ssh0321y\}@mail.ustc.edu.cn}
\\
  \texttt{\{kingdon, chensw, jianyang, xbtan\}@ustc.edu.cn}
}

\begin{document}
\maketitle
\renewcommand{\thefootnote}{\fnsymbol{footnote}}
\footnotetext[1]{Corresponding authors}
\begin{abstract}

Large Language models (LLMs) have demonstrated significant potential in text-to-SQL reasoning tasks, yet a substantial performance gap persists between existing open-source models and their closed-source counterparts. In this paper, we introduce \textbf{SQLForge}, a novel approach for synthesizing reliable and diverse data to enhance text-to-SQL reasoning in LLMs. We improve data reliability through SQL syntax constraints and SQL-to-question reverse translation, ensuring data logic at both structural and semantic levels. We also propose an SQL template enrichment and iterative data domain exploration mechanism to boost data diversity. Building on the augmented data, we fine-tune a variety of open-source models with different architectures and parameter sizes, resulting in a family of models termed \textbf{SQLForge-LM}. SQLForge-LM achieves the state-of-the-art performance on the widely recognized Spider and BIRD benchmarks among the open-source models. Specifically, SQLForge-LM achieves EX accuracy of $85.7\%$ on Spider Dev and $59.8\%$ on BIRD Dev, significantly narrowing the performance gap with closed-source methods.

\end{abstract}

\section{Introduction}

\begin{figure*}[!t]
  \includegraphics[width=\linewidth]{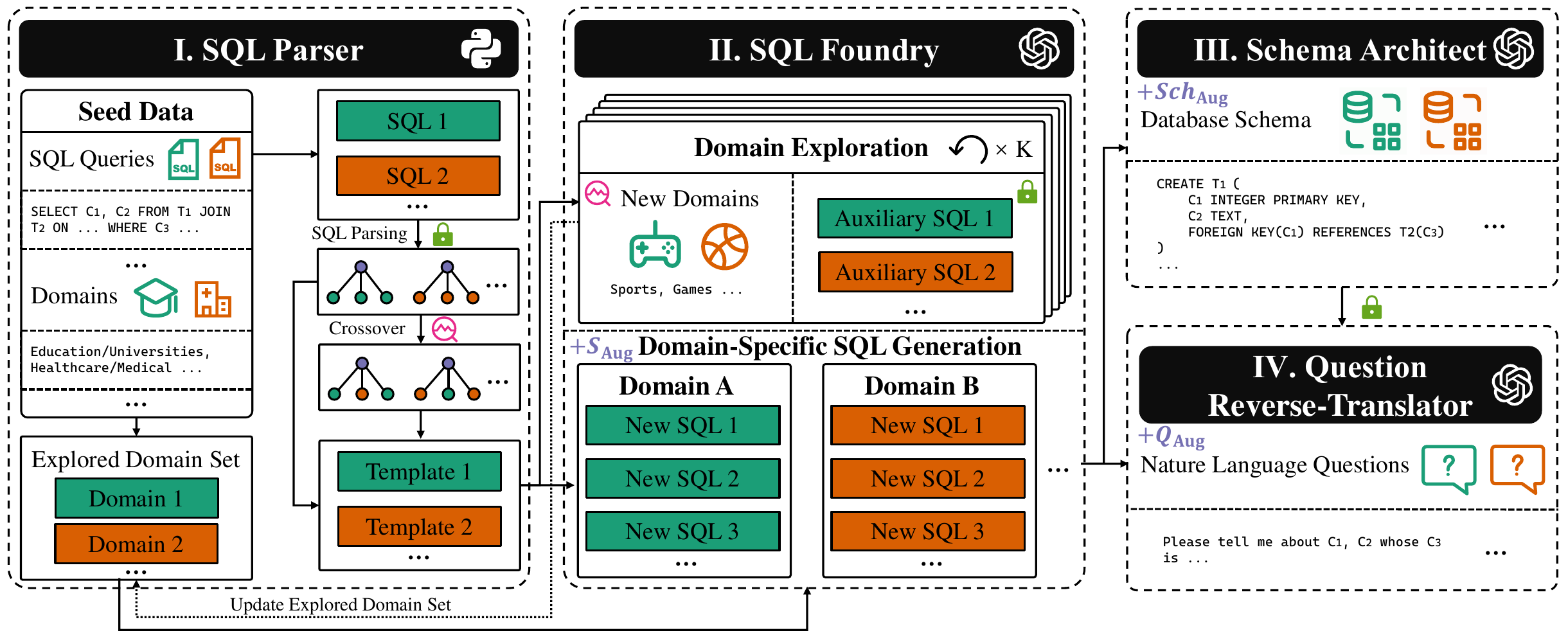}
  \caption{Overview of the proposed SQLForge framework. SQLForge comprises four key components. The SQL Parser generates and enriches SQL templates from seed SQL queries. SQL Foundry generates SQL statements across diverse domains using templates, and upon reaching a domain threshold, it focuses on generating statements within these domains. Then, Schema Architect adds detailed schemas to the generated SQL statements. Finally, Question Reverse-Translator converts the SQL queries into natural language questions aligned with their schemas. The symbol \includegraphics[width=0.3cm]{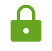} represents reliability enhancements, while \includegraphics[width=0.3cm]{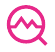} signifies diversity enhancements.}
  \label{fig:1}
\end{figure*}

Text-to-SQL, which transforms natural language questions into SQL queries, serves as a critical bridge between non-expert users and database systems, significantly lowering the barriers to data access \citep{liu2024survey,shi2024survey}. Recently, LLMs have demonstrated exceptional capabilities in various NLP tasks, marking a new paradigm for text-to-SQL solutions. Existing LLM-based approaches \citep{pourreza2024din,gao2023text} usually rely on powerful closed-source models combined with prompt engineering \citep{wei2022chain}. Although these methods achieve impressive performance, they pose challenges such as high computational costs, limited customizability, and significant data privacy concerns.

In response, open-source LLMs have gained attraction due to their lower costs, enhanced data privacy, and greater customizability, making them well-suited for resource-constrained or privacy-sensitive applications. Despite their success in various NLP tasks, open-source models still have a significant performance gap in text-to-SQL \citep{yang2024synthesizing}. For example, the performance difference between CodeLlama-13B and GPT-4 on BIRD benchmark exceeds ${20\%}$. To narrow this gap, we need to synthesize large-scale text-to-SQL data to fine-tune open-source LLMs for enhancing their capabilities for text-to-SQL reasoning.

Existing data synthesis methods for text-to-SQL, such as AnnotatedTables \citep{hu2024annotatedtables} and MultiSQL \citep{li2024multisql}, primarily rely on simple prompt engineering to expand data from existing domains directly. The data generated by these methods suffer from data domain scarcity and poor SQL structure diversity. Other methods like SENSE \citep{yang2024synthesizing}, generate data in new domains directly based on prompting. However, the generated data exhibit poor execution rates and weak semantic alignment between question and SQL pairs, requiring significant time and labor for filtering and refinement. Specifically, according to our evaluation, when the token length is $50$, the SQL statements generated by GPT-4 have a high execution failure rate of about ${14\%}$. To generate text-to-SQL data at scale, we have to overcome two critical challenges: (1) how to ensure the diversity of synthetic data to enhance model generalization, and (2) how to ensure the reliability of synthetic data without manual annotation.

To tackle these challenges, we seek to use syntactic constraints to improve augmented data's reliability, and introduce SQL structure enrichment and domain exploration to expand diversity. Specifically, to guarantee the executability of generated SQL statements, we use the templates derived from parsing valid SQL statements as  the syntactic constraints of SQL for data synthesis. Additionally, to increase the diversity of the synthetic data, we expand the sentence patterns of these templates and propose an iterative domain exploration to generate SQL of entirely new domains. Different from previous work \citep{hu2023importance, kobayashi2024you}, which performs direct SQL-to-question in existing domains, we synthesize data of entirely new domains, generate corresponding database schemas, and incorporate these schemas into the final problem translation, thereby strengthening semantic alignment and enhancing data reliability.

In light of the above idea, we propose a data synthesis framework, named \textbf{SQLForge}, to generate a large-scale, reliable, and diverse text-to-SQL dataset as shown in Fig.\ref{fig:1}. To evaluate SQLForge's effectiveness, we fine-tune several popular open-source pretrained models, such as CodeLlama \citep{roziere2023code}, resulting in a new family of text-to-SQL models named \textbf{SQLForge-LM}. Our experiments demonstrate that SQLForge-LM achieves high performance on well-known benchmarks like Spider \citep{yu2018spider} and BIRD \citep{li2024can}. Specifically, it attains state-of-the-art (SOTA) performance among existing open-source model-based  methods, significantly narrowing the gap with the closed-source model-based methods. Additionally, we evaluate the robustness of SQLForge-LM using datasets meticulously crafted to assess its resistance against perturbations and generalization, including SYN \citep{gan2021towards}, REALISTIC \citep{deng2021structure}, and DK \citep{gan2021exploring}, where it demonstrates strong and consistent performance. Furthermore, we evaluate SQLForge's data synthesis capabilities under corner cases, by employing highly intricate SQL statements, to verify its consistent generation of reliable and high-quality data in these challenging edge scenarios. We also deploy and test our data synthesis framework on open-source models, demonstrating its adaptability in different computational environments.

\begin{figure*}[!t]
  \includegraphics[width=\linewidth]{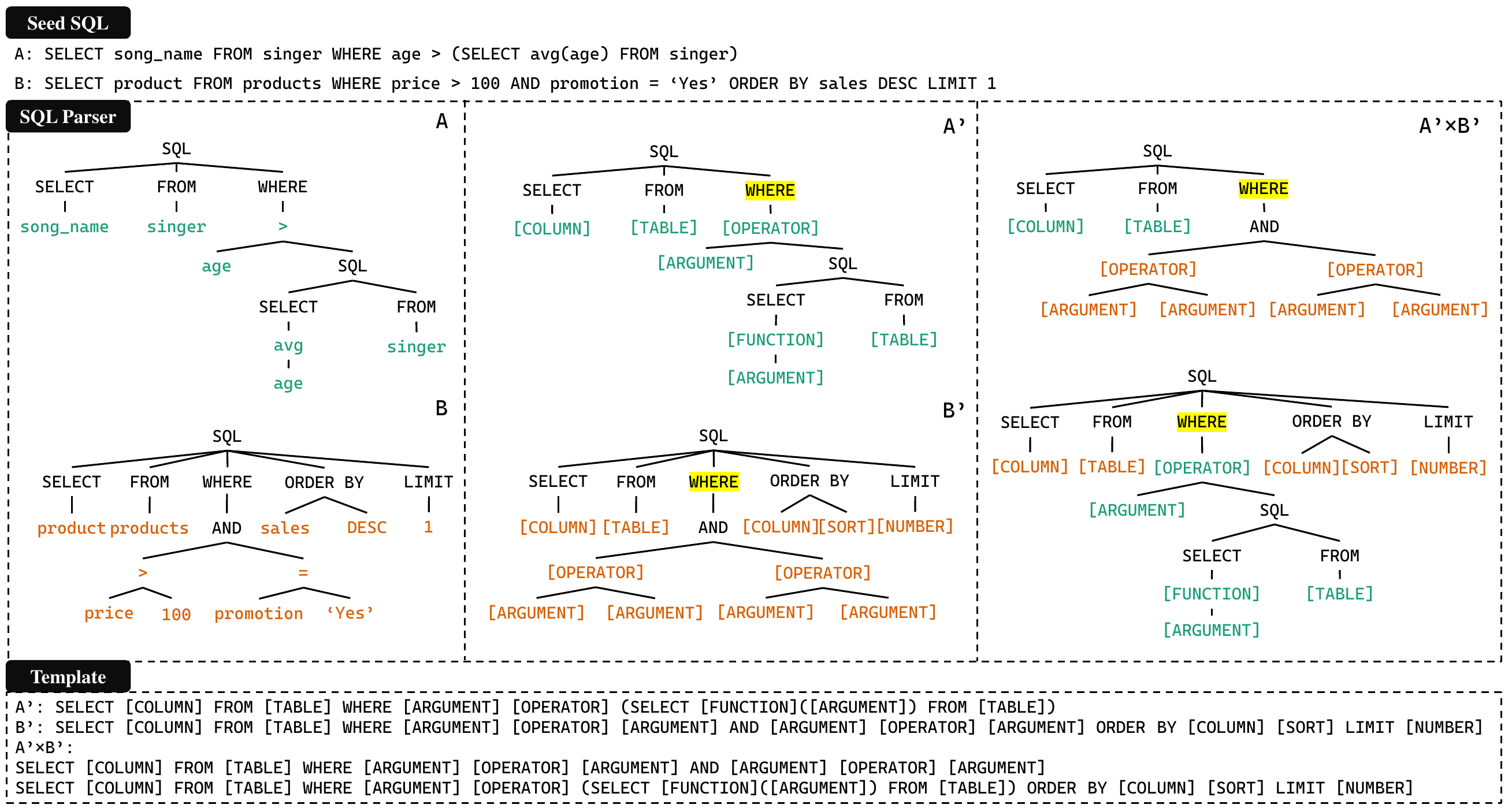}
  \caption{An example of the SQL Parser, which converts SQL into template or generates new template with crossover via AST.}
  \label{fig:2}
\end{figure*}

The main contributions of this paper are summarized as follows:
\begin{itemize}
    \item We propose SQLForge, a framework that incorporates syntax constraints and domain exploration, to drive LLMs to generate reliable and diverse text-to-SQL data.
    \item We fine-tune a new family of open-source models using the data synthesized by SQLForge, i.e. SQLForge-LM, which achieves SOTA performance across multiple benchmarks among methods based on open-source models.
    \item Further evaluation shows that SQLForge is capable of handling extreme data generation tasks and achieving the excellent performance even with open-source base models, significantly enhancing its practical applicability.
\end{itemize}

\section{SQLForge}

In this section, we present SQLForge, a unified framework designed for generating reliable and diverse text-to-SQL data, which consists of four core components: \textit{SQL Parser}, \textit{SQL Foundry}, \textit{Schema Architect}, and \textit{Question Reverse-Translator}, as illustrated in Fig.\ref{fig:1}. Unlike existing data synthesis approaches \citep{li2024multisql, yang2024synthesizing}, which directly generate SQL statements from natural language questions, SQLForge reverses such synthesis process. Initially, the SQL Parser extracts SQL templates from seed data as the inherent syntactic structure. Then, the SQL Foundry leverages these templates to generate SQL statements in new domains. Afterwards, the Schema Architect supplements the database schema according to SQL. Finally, the Question Reverse-Translator transforms the SQL into corresponding natural language questions via reverse translation.

\paragraph{Seed Data} A limited set of seed text-to-SQL data provides the initial corpus for data synthesis. In this paper, the seed data are sourced from the well-known Spider and BIRD datasets, comprising approximately 18K text-to-SQL samples extracted from their training sets. We define this data set as $\mathcal{D}_\text{Seed}=\{(q^i,d^i,{sch}^i,s^i)\}_{i=1}^n$, where $q^i$ represents the $i$-th natural language question, $d^i$ denotes its domain, ${sch}^i$ specifies its database schema, $s^i$ corresponds to its SQL statement answer, and $n$ is the total number of samples. The domains from seed data set are added to the explored domain set $E$, while the SQL data, referred to as the seed SQL, are represented as $S_\text{Seed}$.

\paragraph{SQL Parser} SQL Parser transforms SQL into a structured abstract syntax tree (AST) to standardize the SQL queries while preserving their inherent structure, which converts the seed SQL $S_\text{Seed}$ into the template $T$. Different from the method \citep{kobayashi2024you}, which only replaces the columns and table names in SQL statements, the SQL Parser traverses the tree and replaces non-keyword nodes with placeholders. Each placeholder is annotated with a fill type, which describes the role and context of the original node within the query. Additionally, the parser performs crossover operations on sub-trees with the same keywords for enriching SQL structure. In this process, the SQL templates remain the intact grammatical structure of SQL statements, thereby guaranteeing that the synthetic statement is compliant with the SQL specifications throughout the following synthesis process. A working example of the SQL Parser is illustrated in Fig.\ref{fig:2}.

\paragraph{SQL Foundry} SQL Foundry instantiates templates $T$ to generate diverse SQL statements $S_\text{Aug}$ across new domains, which consists of two steps.
In the first step, to enrich the domain diversity of generated data, we devise an iterative exploration mechanism that dynamically synthesizes new domain names which are specified to the semantic contexts in database scenario. Specifically, SQL Foundry iteratively generates new domain name in conjunction with auxiliary SQL statement tailored to the domain. The generated domain names are ensured to be outside of the explored domain set $E$ mentioned in the Seed Data section. By coupling domain name generation with auxiliary SQL construction, we impose a conditional entropy constraint $H(\alpha|\beta)$ (where $\alpha$ denotes domain name and $\beta$ denotes auxiliary SQL) that compresses the domain naming space into a database scenario compatible distribution subspace, thereby systematically eliminating domain drift risks. Following each iteration, the explored domain set is updated with the newly created domain name.

The second step is triggered when exploring enough domains after $K$ rounds of iteration. At this step, SQL Foundry dynamically binds an SQL template with a domain name derived from $E$, and generates syntactically diverse and domain-specific SQL statements $S_\text{Aug}$, which incorporates both the structural diversity and execution reliability of generated SQL statements in previously unexplored domains.

\paragraph{Schema Architect} Since the newly generated SQL statements belong to entirely new domains, the original database schema is no longer applicable. Schema Architect generates the corresponding database schema expressions $Sch_\text{Aug}$ for the newly generated SQL statements $S_\text{Aug}$. Given that SQL statements are highly structured and free of invalid or redundant information, extracting fields and parsing them into a database schema is highly reliable. Furthermore, constraints such as foreign keys, which are explicitly defined in the SQL statements, can be easily incorporated into the corresponding database schema. The above process ultimately produces the augmented schema $Sch_\text{Aug}$.

\paragraph{Question Reverse-Translator} Question Rever-se-Translator transforms the augmented SQL statements $S_\text{Aug}$ into their corresponding natural language questions $Q_\text{Aug}$. The previous reverse translation approaches \citep{hu2023importance,kobayashi2024you} disregard essential database schema information, resulting in ambiguous entity references and structural misinterpretations. Besides, their over-reliance on SQL patterns often yields syntactically valid but semantically inconsistent questions that fail to reflect authentic human query patterns. To addresses these issues, Question Reverse-Translator integrates the associated database schema $Sch_\text{Aug}$. This design enhances the model's understanding of database topology and entity relationships, enabling alignment between SQL operations and their natural language expressions. The resulting natural language questions $Q_\text{Aug}$ exhibit improved semantic fidelity and linguistic naturalness.

\begin{table}[!t]
  \tiny
  \centering
  \begin{tabularx}{0.5\textwidth}{X}
    \hline
    \textbf{Example of Augmented Data}\\
    \hline
    \#\#\#~Question:\\
    \textcolor{teal}{Which are the top 10 campaigns with the highest total number of clicks?}\\
    \#\#\#~Schema:\\
    \textcolor{blue}{CREATE TABLE} \textcolor{violet}{Campaigns}(\\
    \ \ \ \ \textcolor{violet}{CampaignID} \textcolor{blue}{INTEGER PRIMARY KEY},\\
    \ \ \ \ \textcolor{violet}{CampaignName} TEXT\\
    \ \ \ \ \textcolor{blue}{FOREIGN KEY}(CampaignID) \textcolor{blue}{REFERENCES} \textcolor{violet}{Impressions}(CampaignID)\\
    );\\
    \textcolor{blue}{CREATE TABLE} \textcolor{violet}{Impressions}(\\
    \ \ \ \ \textcolor{violet}{ImpressionID} \textcolor{blue}{INTEGER PRIMARY KEY},\\
    \ \ \ \ \textcolor{violet}{CampaignID} \textcolor{blue}{INTEGER},\\
    \ \ \ \ \textcolor{violet}{Clicks} \textcolor{blue}{INTEGER},\\
    \ \ \ \ \textcolor{blue}{FOREIGN KEY}(CampaignID) \textcolor{blue}{REFERENCES} \textcolor{violet}{Campaigns}(CampaignID)\\
    );\\
    \#\#\#~SQL:\\
    \textcolor{blue}{SELECT} C.CampaignName \textcolor{blue}{FROM} Campaigns \textcolor{blue}{AS} C \textcolor{blue}{JOIN} Impressions \textcolor{blue}{AS} I \textcolor{blue}{ON} C.CampaignID = I.CampaignID \textcolor{blue}{GROUP BY} C.CampaignName \textcolor{blue}{ORDER BY} \textcolor{purple}{SUM}(I.Clicks) \textcolor{blue}{DESC} \textcolor{blue}{LIMIT} \textcolor{violet}{10}\\
    \hline
  \end{tabularx}
  \caption{\label{Data}
    An example of Augmented Data, schema provided in the form of CREATE TABLE statements.
  }
\end{table}

We deployed the described pipeline using GPT-4 as the base model, generating a total of 25K text-to-SQL data points in over 1,000 domains, collectively referred to as Augmented Data ($Q_\text{Aug}, Sch_\text{Aug}, S_\text{Aug}$), an example of Augmented Data is presented in Tab.\ref{Data}. By integrating this augmented data with the Spider and BIRD training sets, we constructed a comprehensive dataset $\mathcal{D}_\text{Aug}$, which served as the training set for fine-tuning open-source pre-trained models, such as CodeLlama. For all $(q^i, sch^{i}, s^i) \in \mathcal{D}_\text{Aug}$, the log likelihood loss function of fine-tuning is defined as: 
\begin{equation}
\scalebox{0.8}{
$\mathbb{E}_{(q^i,sch^{i},s^i)\sim\mathcal{D}_\text{Aug}}\left[\displaystyle\sum_{l=1}^{L}\log p_{\theta}(s_l^i|s_{1:l-1}^i,q^i,sch^{i})\right]$}
\label{equation}
\end{equation}
where $\theta$ represents the model parameters, and $L$ is the length of SQL statement. The resulting text-to-SQL reasoning models are named SQLForge-LM.

\section{Experiments}

\subsection{Experimental Setup}

\paragraph{Datasets} We conduct evaluations on two well-known datasets: Spider \citep{yu2018spider} and BIRD \citep{li2024can}. Additionally, we also evaluate our models on three robust datasets: SYN \citep{gan2021towards}, REALISTIC \citep{deng2021structure}, and DK \citep{gan2021exploring}, to evaluate the robustness and generalization ability of our proposed method. More information about datasets is shown in Appendix~\ref{appendix:B.1}.

\paragraph{Evaluation Metric} We adopt EX (Execution Accuracy, \citealp{yu2018spider}) as the evaluation metric, which measures whether the SQL execution result exactly matches the execution result of provided golden SQL.

\paragraph{Models} We perform LoRA \citep{hu2022lora} fine-tuning on two model families, including CodeLlama 7B and 13B \citep{roziere2023code}, Qwen2 0.5B and 7B \citep{yang2024qwen2}. Fine-tuning details are described in Appendix~\ref{appendix:B.2}.

\paragraph{Compared Methods} We compare our approach against four categories of baselines:

\subparagraph{Closed-Source Models} CodeX \citep{chen2021evaluating}, ChatGPT \citep{ouyang2022training}, PaLM \citep{anil2023palm}, GPT-4 \citep{achiam2023gpt}, and Claude-2 \citep{anthropic2023introducing}.

\subparagraph{Open-Source Models} Qwen-2 \citep{yang2024qwen2}, DeepSeek-Coder \citep{guo2024deepseek}, CodeLlama \citep{roziere2023code}, Gemma \citep{team2024gemma}, StarCoder \citep{lozhkov2024starcoder}, and Llama-3 \citep{dubey2024llama}.

\subparagraph{Prompting Closed-Source Models} CHASE-SQL \citep{pourreza2024chase}, DAIL-SQL \citep{gao2023text}, DIN-SQL \citep{pourreza2024din}, PTD-SQL \citep{luo2024ptd}, and TA-SQL \citep{qu2024before}.

\subparagraph{Fine-tuning Open-Source Models} CodeS \citep{li2024codes}, PICARD \citep{scholak2021picard}, RESDSQL \citep{li2023resdsql}, and SENSE \citep{yang2024synthesizing}.

\begin{table*}[!t]
  \small
  \centering
  \begin{tabularx}{\textwidth}{cCCC|CCC|C}
    \toprule
    \multirow{2}*{\textbf{Model}} & \multirow{2}*{\textbf{Base}} & \multirow{2}*{\textbf{Size}} & \multirow{2}*{\textbf{\#\ Calls}} & \multicolumn{2}{c}{\textit{\textbf{Spider}}} & \textit{\textbf{BIRD}} & \multirow{2}*{\textbf{Average}}\\
    ~ & ~ & ~ & ~ & \textit{\textbf{Dev}} & \textit{\textbf{Test}} & \textit{\textbf{Dev}} & ~\\ \midrule[1pt]
    \multicolumn{8}{c}{\textit{\textbf{Closed-Source Models}}}\\ \midrule[1pt]
    CodeX & - & 175B & \XSolidBrush & 71.8 & - & 34.4 & -\\
    ChatGPT & - & - & \XSolidBrush & 72.3 & - & 37.2 & -\\
    PaLM-2 & - & - & \XSolidBrush & - & - & 27.4 & -\\
    GPT-4 & - & - & \XSolidBrush & 72.9 & - & 46.4 & -\\
    Claude-2 & - & - & \XSolidBrush & - & - & 42.7 & -\\ \midrule[1pt]
    \multicolumn{8}{c}{\textit{\textbf{Open-Source Models}}}\\ \midrule[1pt]
    Qwen-2 & - & 0.5B & \XSolidBrush & 44.8 & 43.1 & 14.5 & 34.1\\
    DeepSeek-Coder & - & 1.3B & \XSolidBrush & 59.7 & 58.4 & 22.5 & 46.9\\
    CodeLlama & - & 7B & \XSolidBrush & 61.9 & 62.6 & 23.5 & 49.3\\ 
    Gemma & - & 7B & \XSolidBrush & 50.1 & 50.4 & 21.2 & 40.6\\
    StarCoder & - & 7B & \XSolidBrush & 61.5 & 62.1 & 21.0 & 48.2\\
    Qwen-2 & - & 7B & \XSolidBrush & 52.6 & 55.7 & 20.1 & 42.8\\
    CodeLlama & - & 13B & \XSolidBrush & 63.5 & 64.4 & 23.9 & 50.6\\ 
    Llama-3 & - & 70B & \XSolidBrush & 67.4 & 68.1 & 30.6 & 55.4\\ \midrule[1pt]
    \multicolumn{8}{c}{\textit{\textbf{Prompting Closed-Source Models}}}\\ \midrule[1pt]
    CHASE-SQL & Gemini-1.5 & - & \CheckmarkBold & - & \textbf{87.6} & \textbf{73.0} & -\\ 
    DAIL-SQL & GPT-4 & - & \CheckmarkBold & 83.5 & 86.6 & 54.8 & 75.0\\ 
    DIN-SQL & GPT-4 & - & \CheckmarkBold & 82.9 & 85.3 & 50.7 & 73.0\\ 
    PTD-SQL & GPT-4 & - & \CheckmarkBold & \textbf{85.7} & - & 57.0 & -\\ 
    TA-SQL & GPT-4 & - & \CheckmarkBold & \underline{85.0} & - & 56.2 & -\\ \midrule[1pt]
    \multicolumn{8}{c}{\textit{\textbf{Fine-tuning Open-Source Models}}}\\ \midrule[1pt]
    PICARD & T5 & 3B & \XSolidBrush & 79.3 & 75.1 & - & -\\
    RESDSQL & RoBERTa & 3B & \CheckmarkBold & 84.1 & 79.9 & - & -\\
    SENSE & CodeLlama & 13B & \XSolidBrush & 84.1 & 86.6 & 55.5 & 75.4\\ 
    CodeS & StarCoder & 15B & \XSolidBrush & 84.9 & - & 58.5 & -\\ \midrule[1pt]
    \multicolumn{8}{c}{\textit{\textbf{Ours}}}\\ \midrule[1pt] \rowcolor{gray!10}
    SQLForge-LM & Qwen2 & 0.5B & \XSolidBrush & 76.1 & 75.3 & 36.9 & 62.8\\ \rowcolor{gray!10}
    SQLForge-LM & Qwen2 & 7B & \XSolidBrush & 82.5 & 82.9 & 54.1 & 73.2\\ \rowcolor{gray!10}
    SQLForge-LM & CodeLlama & 7B & \XSolidBrush & 84.4 & 85.8 & 56.9 & \underline{75.7}\\ \rowcolor{gray!10}
    SQLForge-LM & CodeLlama & 13B & \XSolidBrush & \textbf{85.7} & \underline{87.4} & \underline{59.8} & \textbf{77.6}\\ \midrule[1pt]
  \end{tabularx}
  \caption{\label{Spidr-BIRD}
    Performance comparison on Spider and BIRD. \textbf{\# Calls} indicates whether the method needs to call the model multiple times.
  }
\end{table*}

\subsection{Main Results}

Tab.\ref{Spidr-BIRD} presents the EX accuracy of SQLForge-LMs on Spider and BIRD, the results reveal the following observations: (1) Among open-source models, SQLForge-LM achieves state-of-the-art (SOTA) EX accuracy, SQLForge-LM (CodeLlama-13B variant) is $2.2\%$ ahead of SENSE-13B in average performance, even when SQLForge-LM uses the LoRA fine-tuning and SENSE uses full parameter fine-tuning. (2) SQLForge-LM demonstrates high performance across models with varying parameter scales. In the case of CodeLlama, the average performance of the 7B and 13B models increased by $26.4\%$ and $27.0\%$ respectively, showcasing strong parametric scalability. (3) The model consistently performs well across different model families, highlighting its transferability. (4) SQLForge-LM not only surpasses many closed-source models but also reduces the gap between open-source model and prompting closed-source approaches. SQLForge-LM achieves performs comparable to most prompting closed-source approaches, although a significant performance gap remains with CHASE-SQL on BIRD. The performance reported in Tab.\ref{Spidr-BIRD} are all obtained using greedy decoding.

Additionally, as shown in Tab.\ref{ROBUST}, we evaluate the performance of SQLForge-LM on the Robust datasets (SYN, REALISTIC, DK). SQLForge-LM secures a leading position across these datasets without extra training. Using the same base model, SQLForge-LM-7B and SQLForge-LM-13B outperform SENSE-7B and SENSE-13B by an average of $1.1\%$ and $1.3\%$, respectively, showcasing strong resilience to perturbations and exceptional generalization capabilities.

\begin{table*}[!t]
  \small
  \centering
  \begin{tabularx}{\textwidth}{ccC|CCC|C}
    \toprule
    \textbf{Model} & \textbf{Base} & \textbf{\#\ Calls} & \textit{\textbf{SYN}} & \textit{\textbf{REALISTIC}} & \textit{\textbf{DK}} & \textbf{Average} \\ \midrule[1pt]
    RESDSQL & RoBERTa-3B & \CheckmarkBold & 76.9 & 81.9 & 66.0 & 74.9\\
    SENSE & CodeLlama-7B & \XSolidBrush & 72.6 & 82.7 & 77.9 & 77.7\\
    SENSE & CodeLlama-13B & \XSolidBrush & \underline{77.6} & \underline{84.1} & \underline{80.2} & \underline{80.6}\\
    CodeS & StarCoder-15B & \XSolidBrush & 77.0 & 83.1 & 70.7 & 76.9\\ \midrule[1pt] \rowcolor{gray!10}
    SQLForge-LM & CodeLlama-7B & \XSolidBrush & 74.6 & 83.3 & 78.9 & 78.8\\ \rowcolor{gray!10}
    SQLForge-LM & CodeLlama-13B & \XSolidBrush & \textbf{78.9} & \textbf{84.8} & \textbf{82.1} & \textbf{81.9}\\ \midrule[1pt]
  \end{tabularx}
  \caption{\label{ROBUST}
    Performance comparison on robust datasets. \textbf{\# Calls} indicates whether the method needs to call the model multiple times.
  }
\end{table*}

\subsection{Ablation Study}

The following section details the ablation studies conducted in this work, all of which utilize CodeLlama-7B as the base model.

\paragraph{Analysis of Data Composition} We investigate the impact of different data components on model performance by training the model with various combinations of datasets, as shown in Tab.\ref{different-data}. When using only the Spider training set, the model's performance on Spider improves significantly ($15.0\%$). Similarly, using the BIRD training set leads to notable improvements on BIRD ($25.6\%$). These results are consistent with the inherent characteristics of the datasets. Compared to Spider, BIRD is a more complex dataset from a distinct domain. Notably, data from disjoint domains, such as BIRD and Spider, can still complement each other ($6.7\%$ on Spider, $13.1\%$ on BIRD), indicating that diverse domain data can stimulate the model's implicit domain adaptability. Because our augmented data not only cover both simple and complex SQL generation tasks but also expands the domain scope substantially. When adding augmented data, the model's performance is significantly improved on both datasets ($22.5\%$ on Spider, $33.4\%$ on BIRD), demonstrating the effectiveness of our data augmentation strategy.

\begin{table}[!t]
  \centering
  \resizebox{\linewidth}{!}{
  \begin{tabular}{ccc|cc|c}
    \toprule
    \multicolumn{3}{c|}{\textbf{Data}} & \textbf{\textit{Spider}} & \textbf{\textit{BIRD}} & \multirow{2}*{\textbf{Average}}\\
    \textbf{\textit{Spider}} & \textbf{\textit{BIRD}} & \textbf{\textit{Augmented~Data}} & \textbf{\textit{Dev}} & \textbf{\textit{Dev}}\\ \midrule[1pt]
    \CheckmarkBold & \XSolidBrush & \XSolidBrush & 76.9 \cellcolor{gray!20} & 36.6 \cellcolor{gray!10} & 56.8 \cellcolor{gray!10} \\
    \XSolidBrush & \CheckmarkBold & \XSolidBrush & 68.6 \cellcolor{gray!10} & 49.1 \cellcolor{gray!20} & 58.9 \cellcolor{gray!20} \\
    \CheckmarkBold & \CheckmarkBold & \XSolidBrush & 79.8 \cellcolor{gray!40} & 50.5 \cellcolor{gray!40} & 65.2 \cellcolor{gray!40} \\
    \CheckmarkBold & \CheckmarkBold & \CheckmarkBold & 84.4 \cellcolor{gray!80} & 56.9 \cellcolor{gray!80} & 70.7 \cellcolor{gray!80} \\ \midrule[1pt]
  \end{tabular}
  }
  \caption{\label{different-data}
    Effect of different data composition with CodeLlama-7B as the base model.
  }
\end{table}

\begin{figure}[!t]
  \includegraphics[width=0.9\linewidth]{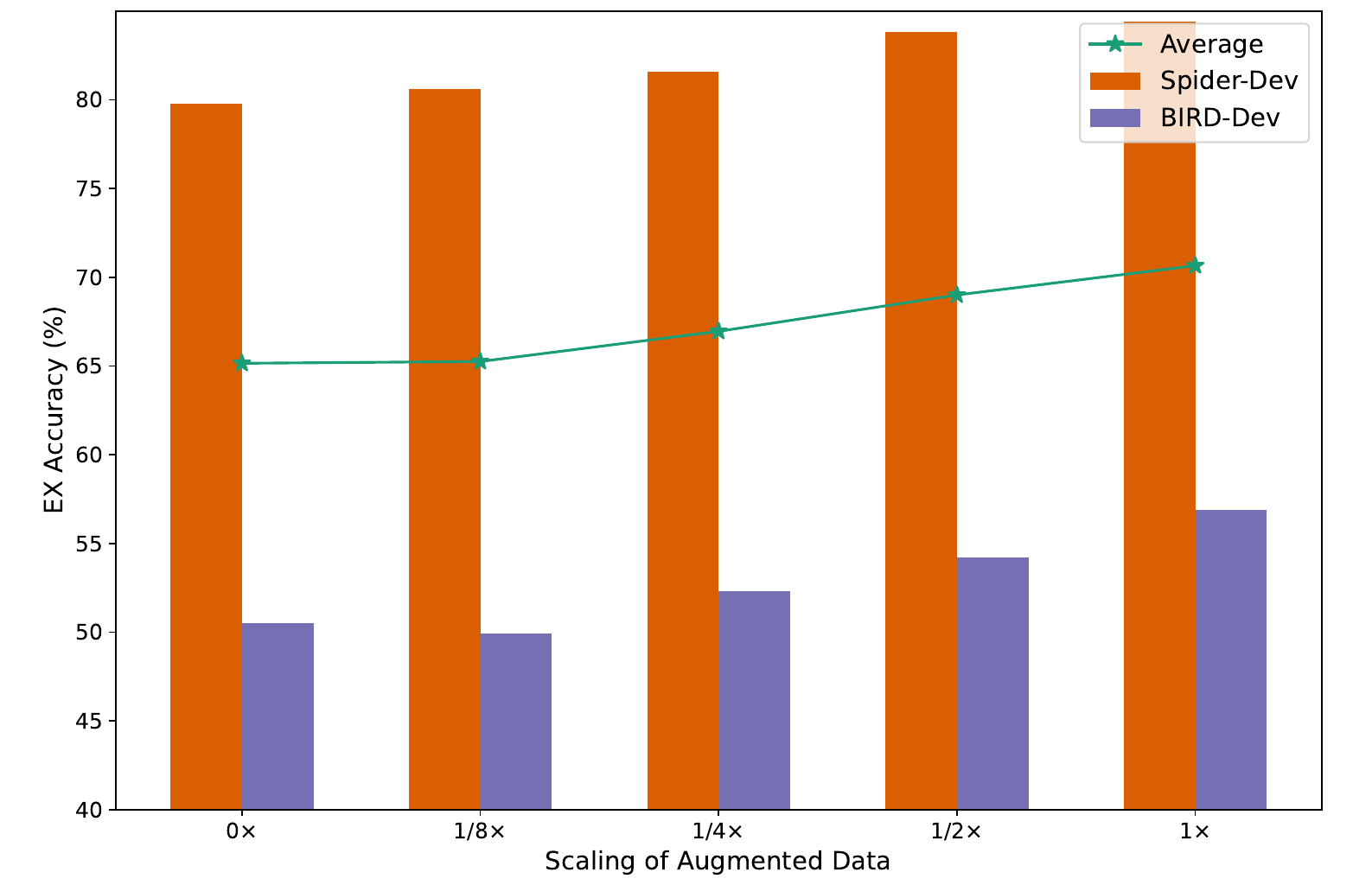}
  \caption{Effect of different scaling of augmented data with CodeLlama-7B as the base model.}
  \label{fig:3}
\end{figure}

\paragraph{Analysis of Scaling of Augmented Data} We analyze the scaling performance of the augmented data. Specifically, we train the model using $0\times$, $1/8\times$, $1/4\times$, $1/2\times$ and $1\times$ augmented data. The results are shown in Fig.\ref{fig:3}. As the amount of augmented data increases, the model's performance consistently improves, demonstrating excellent scaling performance of our augmented data.

\paragraph{Analysis of Data Augmentation methods} We compare our data augmentation approach with direct data augmentation \citep{yang2024synthesizing} and evaluate the impact of our component designs. Direct data augmentation generates new data following the sequence from schema to question, and ultimately to SQL. Additionally, we assess the effects of auxiliary SQL and schema enhancement. As shown in Tab.\ref{Augmentation}, our method outperforms existing data synthesis processes, and both auxiliary SQL and schema enhancements contribute to performance improvements.

\paragraph{Analysis of Augmented Data Diversity}

\begin{figure}[!t]
  \centering
  \includegraphics[width=0.5\linewidth]{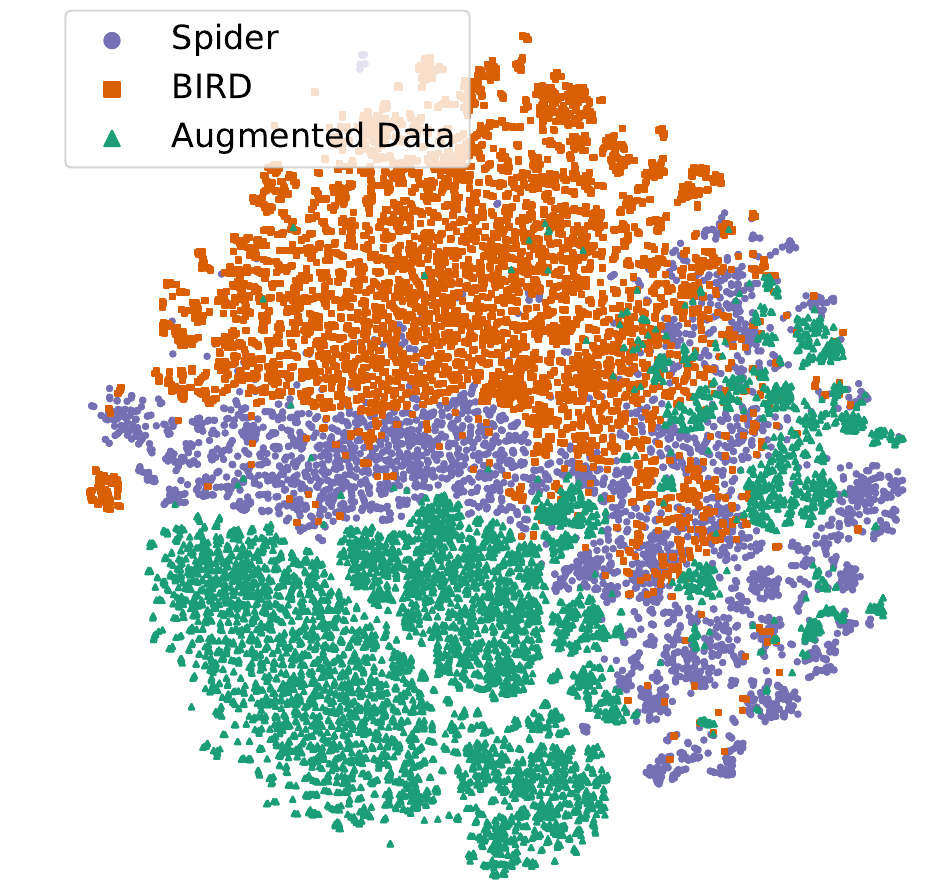}
  \caption{2-D t-SNE illustrates the distribution of seed and augmented data in the semantic space.}
  \label{fig:data}
\end{figure}

\begin{figure}[!t]
  \centering
  \includegraphics[width=0.55\linewidth]{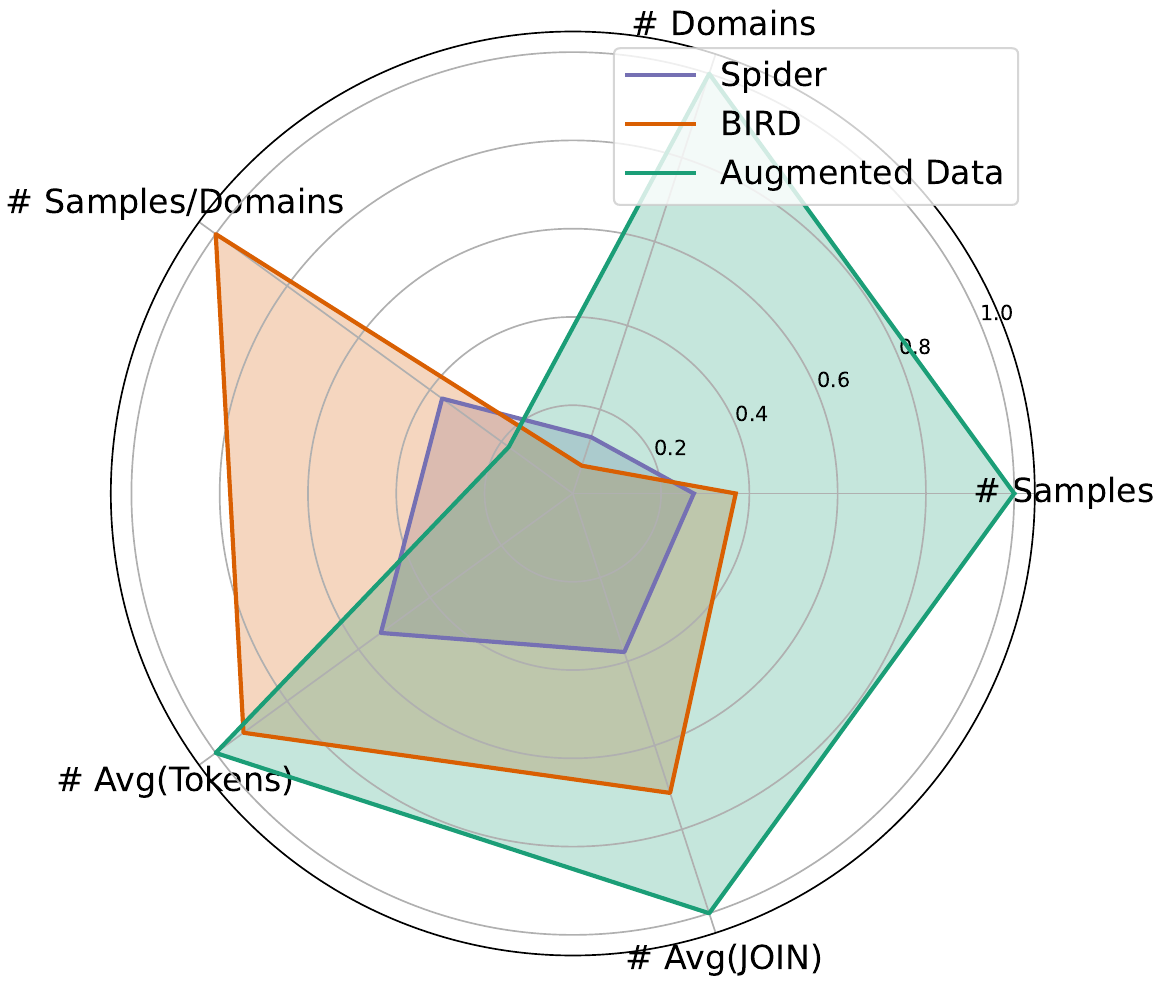}
  \caption{Detailed comparison between seed data and augmented data.}
  \label{fig:6}
\end{figure}

We utilize the T5 model \citep{raffel2020exploring} to generate embedding vectors from the Spider training set, BIRD training set, and augmented data, which are then projected into a two-dimensional space using t-SNE. As shown in Fig.\ref{fig:data}, our augmented data effectively fills and expands the distribution within the semantic space. Additional details on the augmented data can be found in Fig.\ref{fig:6}.

\subsection{Robustness Study}

We evaluate the robustness of SQLForge by examining its scalability and adaptability. Scalability measures SQLForge's performance with larger or more complex inputs, while adaptability measures its stability on open-source models. This analysis demonstrates SQLForge's potential in both complex real-world scenarios and high privacy-sensitive scenarios.

\begin{table}[!t]
  \centering
  \resizebox{\linewidth}{!}{
  \begin{tabular}{l|cc|c}
    \toprule
    \multirow{2}*{\textbf{Augmentation Method}} & \textbf{\textit{Spider}} & \textbf{\textit{BIRD}} & \multirow{2}*{\textbf{Average}}\\
    ~ & \textbf{\textit{Dev}} & \textbf{\textit{Dev}}\\ \midrule[1pt]
    \textbf{Direct Data Augmentation} & 82.9 & 52.5 & 67.7\tiny \textcolor{red}{(-3.0)}\\
    \textbf{SQLForge w/o Auxiliary SQL} & 84.2 & 55.8 & 70.0\tiny \textcolor{red}{(-0.7)}\\
    \textbf{SQLForge w/o Schema} & 83.8 & 53.9 & 68.9\tiny \textcolor{red}{(-1.8)}\\
    \textbf{SQLForge (Ours)} & 84.4 & 56.9 & 70.7\\ \midrule[1pt]
  \end{tabular}
  }
  \caption{\label{Augmentation}
    Comparison of different data augmentation methods. \textbf{SQLForge w/o Auxiliary SQL} presents don't generate auxiliary SQL in Domain Exploration stage. \textbf{SQLForge w/o Schema} presents don't use schema to Enhance inputs.
  }
\end{table}

\begin{figure}[!t]
  \includegraphics[width=0.9\linewidth]{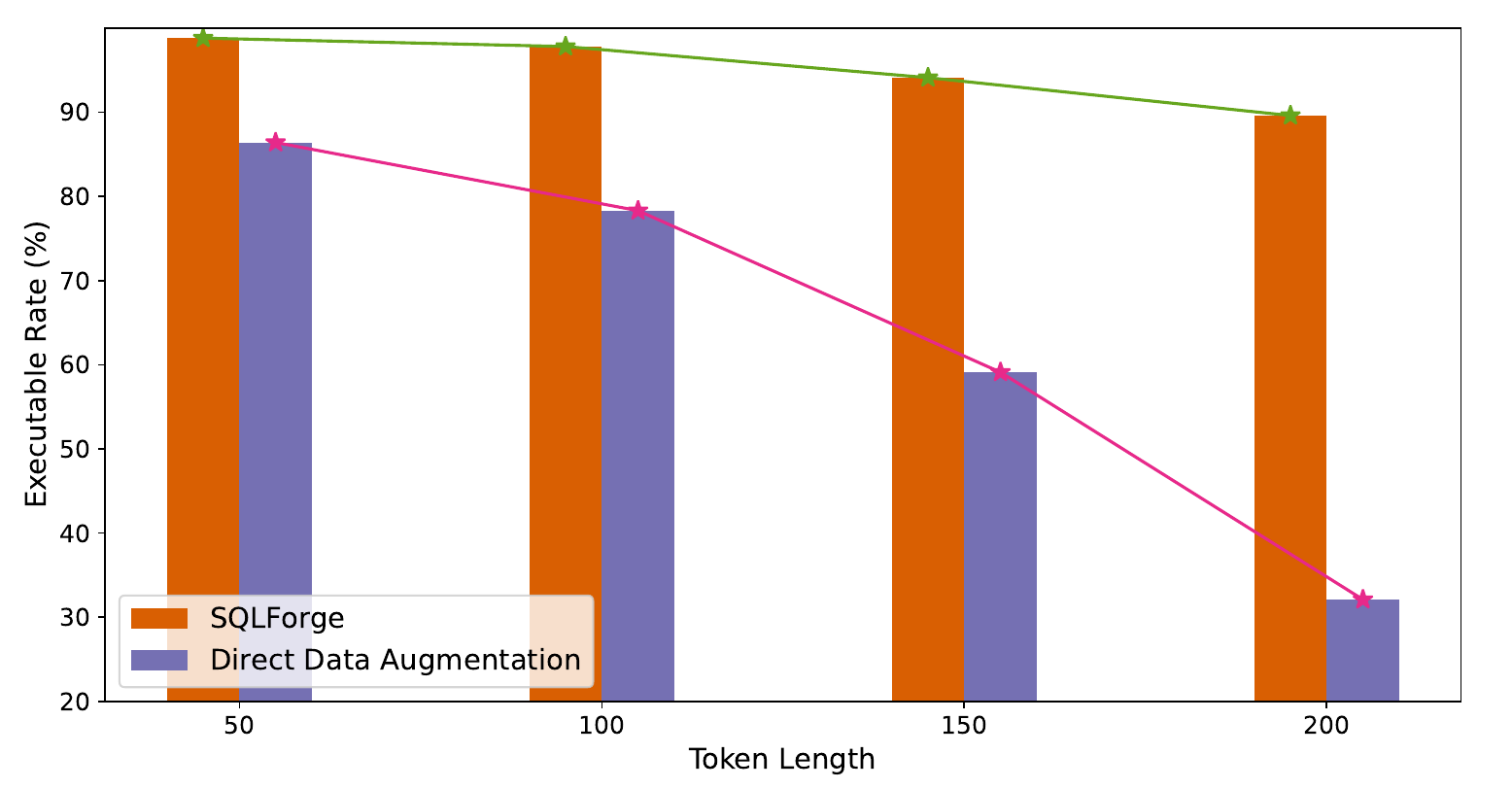}
  \caption{Executable rate of augmented data under different complexity.}
  \label{fig:4}
\end{figure}

\paragraph{Scalability} We employ SQLForge and direct data augmentation approach to produce text-to-SQL data of varying complexities. The complexity of SQL is quantified by token length. We use GPT-4 to generate 1K data samples for each token length category ($50$, $100$, $150$, $200$), and subsequently assess the executable rate of the generated data. The results are illustrated in Fig.\ref{fig:4}. As the complexity of the generated data progressively increases, the executable rate of the data produced through direct data augmentation experiences a significant decline, whereas the rate for data generated by SQLForge exhibits only a marginal decrease.

\begin{figure}[!t]
  \includegraphics[width=0.95\linewidth]{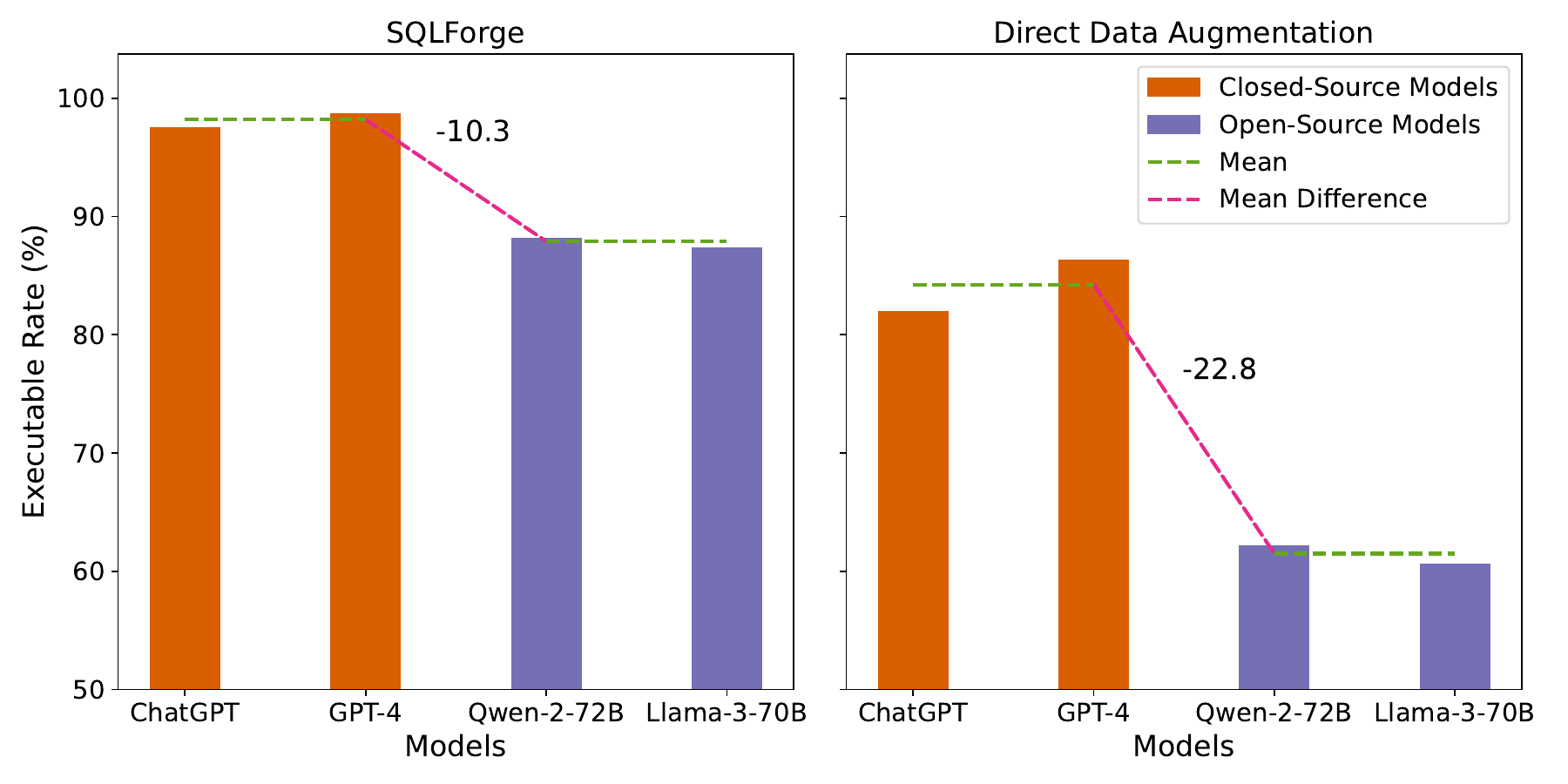}
  \caption{Executable rate of augmented data for open-source and closed-source models.}
  \label{fig:5}
\end{figure}

\paragraph{Adaptability} We select Llama-3-70B, Qwen-2-72B as the open-source base models and ChatGPT, GPT-4 as the closed-source base model. Utilizing both SQLForge and direct data augmentation, we generate text-to-SQL data comprising 1K samples (The token length of SQL is $50$). The results are presented in Fig.\ref{fig:5}. When transitioning to the open-source model, SQLForge maintains stable performance, whereas direct data augmentation exhibits significant fluctuations as the model's capabilities diminish.

\section{Related Work}

\paragraph{Text-to-SQL Reasoning Methods} In the field of text-to-SQL, early approaches like IRNET \citep{guo2019towards} relied on attention-based models to learn intermediate representations. Subsequently, the focus shifted to fine-tuning pre-trained models, with works such as RESDSQL \citep{li2023resdsql}, RAT-SQL \citep{wang2020rat}, and PICARD \citep{scholak2021picard}. The landscape of text-to-SQL research has been further revolutionized by the advent of large language models (LLMs), which have introduced powerful zero-shot and in-context learning capabilities. This advancement has led to the widespread adoption of prompt-based methods using closed-source models. Notable contributions in this area include DAIL-SQL \citep{gao2023text}, which integrated various prompt engineering techniques to enhance performance, DIN-SQL \citep{pourreza2024din}, which decomposed complex tasks into manageable sub-tasks, PTD-SQL \citep{luo2024ptd}, which employed query group partitioning to enable LLMs to focus on specific problem types, TA-SQL \citep{qu2024before}, which utilized task alignment to mitigate hallucinations at each stage of the reasoning process, CHASE-SQL \citep{pourreza2024chase}, which leveraged LLMs' intrinsic knowledge to generate diverse and high-quality SQL candidates using different LLM generators. Additionally, some works have focused on training or fine-tuning LLMs specific to text-to-SQL realm, such as SENSE \citep{yang2024synthesizing} and CodeS \citep{li2024codes}. Our work fine-tunes open-source LLMs through reliable and diverse data augmentation to enhance text-to-SQL reasoning.

\paragraph{Data Augmentation for Text-to-SQL} To increase the quantity of text-to-SQL data, many existing approaches \citep{hu2024annotatedtables,li2024multisql} typically augment data directly based on existing databases. \citet{li2024codes} generates data for specific domain using bi-directional augmentation with manual annotation. \citet{yang2024synthesizing} synthesizes data for new domains through direct data augmentation. In contrast, our work reverses the data augmentation process. By taking advantage of the syntax constraints inherent in the template and employing an iterative domain exploration mechanism, we generate a large volume of reliable and diverse SQL data automatically.

\section{Conclusion}

In this study, we present SQLForge, a comprehensive framework designed to synthesize high-quality data for enhancing text-to-SQL reasoning. Experiment demonstrates that SQLForge-LM achieves SOTA performance in open-source model-based methods on well-known benchmarks, significantly narrowing the performance gap between open-source and closed-source models. Additionally, the framework exhibits robustness and transferability, offering valuable insights into the development of text-to-SQL models and inspiration for other reasoning tasks.

\section*{Limitations}

Computational resource limitations restrict our ability to fine-tune larger models, leaving the performance implications of data synthesis largely unexplored in this context. Additionally, a comparative analysis between full parameter fine-tuning and LoRA fine-tuning is necessary to establish best practices. While our data generation framework has demonstrated strong performance with closed-source models like GPT-4 and shows adaptability with open-source models such as Llama-3-70B, considerations of cost efficiency and privacy drive the need for further exploration of text-to-SQL data generation with open-source models. Unfortunately, resource constraints prevent us from investigating the data generation performance of fine-tuned open-source models.

\section*{Ethics Statement}

Our work aims to provide a low-cost solution for text-to-SQL scenarios by enhancing open-source models using synthetic data. However, like any language model, it may generate unrealistic or even harmful content. We strongly encourage users to review the outputs when utilizing SQLForge and SQLForge-LM. Additionally, our research leverages open-source models such as CodeLlama and Qwen-2, as well as software frameworks like PyTorch and HuggingFace. We adhere to the policies and licenses of these resources and acknowledge their significant contributions to our work.

\section*{Acknowledgement}

This work was financially supported by National Key Research and Development Program of China, No.2024YDLN0004.

\bibliography{custom}

\appendix

\section{Prompt Design}
\label{appendix:A}

\subsection{Prompt of SQL Foundry}
\label{appendix:A.1}

Tab.\ref{Prompt1} presents the prompt format used in Domain Exploration,  Tab.\ref{Prompt2} presents the prompt format used in Domain-Specific SQL Generation.

\subsection{Prompt of Schema Architect}
\label{appendix:A.2}

Tab.\ref{Prompt3} presents the prompt format used in Schema Architect.

\subsection{Prompt of Question Reverse-Translator}
\label{appendix:A.3}

Tab.\ref{Prompt4} presents the prompt format used in Question Reverse-Translator.

\section{Experiment Details}
\label{appendix:B}

\subsection{Datasets Details}
\label{appendix:B.1}

Spider consists of 10,181 questions and 5,693 unique complex SQL queries on 200 databases with multiple tables covering 138 different domains, while BIRD contains over 12,751 unique question-SQL pairs, 95 big databases with a total size of 33.4 GB. It also covers more than 37 professional domains, such as blockchain, hockey, healthcare and education, etc, SYN contains 7,000 training and 1,034 development examples with replacing simple string-matched problem tags or patern names with their synonyms, REALISTIC contains 508 text-to-SQL pairs from Spider, replaced mentioned schema items in questions to make them closer to realworld scenarios, DK contains 535 text-to-SQL pairs drawn from the Spider development set, where 270 pairs are the same as the original Spider samples, while the rest 265 pairs are modified to incorporate the domain knowledge.

\subsection{Fine-tuning Details}
\label{appendix:B.2}

Our experiments are conducted using the HuggingFace library and leverage 6 NVIDIA A100 40GB GPUs. We employ the AdamW optimizer with a learning rate of $2e^{-4}$ and a cosine learning rate scheduler, the warm-up phase accounts for $1\%$ of the total training steps. For the LoRA adaptation, we set the rank $r$ to $128$ and the scaling factor $\alpha$ to $256$, applying LoRA modules to all optional target modules in the models.

\clearpage

\begin{table*}[!t]
  \small
  \centering
  \begin{tabularx}{\textwidth}{X}
    \hline
    \textbf{Prompt of Domain Exploration}\\
    \hline
    Your task is to generate a new domain and SQLite based on the provided template.\\
    Requirements:\\
    1. Domain: Avoid domains listed below and ensure diversity by exploring new or under-represented domains.\\
    \ \ \ \ - Existing domains: \{domain\_explored\}\\
    2. Template: \{template\}\\
    \ \ \ \ - The "[]" placeholders should be filled independently and meaningfully based on their context. These placeholders do not need to be identical, even if they appear multiple times in the template.\\
    \ \ \ \ - The content of each placeholder should align with the intent of the query and the domain.\\
    3. Guidelines:\\
    \ \ \ \ - Strictly adhere to the template structure.\\
    \ \ \ \ - Ensure the SQLite statement is meaningful.\\
    \ \ \ \ - Limit the domain name to \{domain\_count\} word(s).\\
    \ \ \ \ - Refrain from adding unrelated content or remarks.\\
    (examples goes here...)\\
    \hline
  \end{tabularx}
  \caption{\label{Prompt1}
    Prompt of Domain Exploration, "\{domain\_explored\}" is replaced with the explored domain set, "\{template\}" is replaced with the template from $T$, "\{domain\_count\}" is chosen from \{1,2,3\}.
  }
\end{table*}

\begin{table*}[!t]
  \small
  \centering
  \begin{tabularx}{\textwidth}{X}
    \hline
    \textbf{Prompt of Domain-Specific SQL Generation}\\
    \hline
    Your task is to generate a SQLite query based on the following template and domain.\\
    Requirements:\\
    1. Domain: \{domain\}\\
    2. Template: \{template\}\\
    \ \ \ \ - The "[]" placeholders should be filled independently and meaningfully based on their context. These placeholders do not need to be identical, even if they appear multiple times in the template.\\
    \ \ \ \ - The content of each placeholder should align with the intent of the query and the domain.\\
    3. Guidelines:\\
    \ \ \ \ - Strictly adhere to the template structure.\\
    \ \ \ \ - Ensure the SQLite query is meaningful, logically coherent, and specific to the domain.\\
    \ \ \ \ - Provide only the SQLite query without any additional explanations or comments.\\
    (examples goes here...)\\
    \hline
  \end{tabularx}
  \caption{\label{Prompt2}
    Prompt of Domain-Specific SQL Generation, "\{domain\}" is chosen from the explored domain set, "\{template\}" is replaced with the template from $T$.
  }
\end{table*}

\begin{table*}[!t]
  \small
  \centering
  \begin{tabularx}{\textwidth}{X}
    \hline
    \textbf{Prompt of Schema Architect}\\
    \hline
    Your task is to generate a schema (CREATE TABLE statements) based on the given SQLite and domain.\\
    Requirements:\\
    1. Domain:\{domain\}\\
    2. SQLite:\{sql\}\\
    3. Guidelines:\\
    \ \ \ \ - Strictly adhere to the column names and types as implied in the SQLite statement.\\
    \ \ \ \ - Ensure that foreign keys are correctly represented without using ALTER TABLE.\\
    \ \ \ \ - Follow the output format strictly, including table names, column names, types, and foreign key constraints.\\
    \ \ \ \ - Do not add any unrelated content or explanations.\\
    (examples goes here...)\\
    \hline
  \end{tabularx}
  \caption{\label{Prompt3}
    Prompt of Schema Architect, "\{sql\}" is replaced with the SQL statement from $S_\text{Aug}$, "\{domain\}" is its domain.
  }
\end{table*}

\begin{table*}[!t]
  \small
  \centering
  \begin{tabularx}{\textwidth}{X}
    \hline
    \textbf{Prompt of Question Reverse-Translator}\\
    \hline
    Your task is to generate a human-readable, natural language question based on the following SQLite query and the schema.\\
    SQLite: \{sql\}\\
    schema: \{schema\}\\
    Guidelines:\\
    \ \ \ \ - Focus on the query's intent, not just repeating table or column names.\\
    \ \ \ \ - Keep the question clear, concise, and intuitive, following natural language patterns.\\
    \ \ \ \ - For aggregations (e.g., COUNT, SUM), joins, or filters, ask about the result or insight, not the structure.\\
    \ \ \ \ - Ensure the question is contextually relevant to the schema and uses table/column names meaningfully.\\
    \ \ \ \ - Avoid unnecessary content or comments.\\
    (examples goes here...)\\
    \hline
  \end{tabularx}
  \caption{\label{Prompt4}
    Prompt of Question Reverse-Translator, "\{sql\}" is replaced with the SQL statement from $S_\text{Aug}$, "\{schema\}" is its schema.
  }
\end{table*}

\end{document}